\documentclass[11pt]{article}

\usepackage[final]{acl}

\usepackage{times}
\usepackage{latexsym}
\usepackage[T1]{fontenc}
\usepackage[utf8]{inputenc}
\usepackage{microtype}
\usepackage{inconsolata}
\usepackage{graphicx}
\usepackage{booktabs}
\usepackage{multirow}
\usepackage{amsmath}
\usepackage{array}
\usepackage{tikz}
\usetikzlibrary{arrows.meta,positioning,fit}
\usepackage{url}

\newcommand{\shroom}{SHROOM-Visions}
\newcommand{\team}{SKstars}
\newcommand{\qwenlarge}{Qwen2.5-VL-72B-Instruct}
\newcommand{\qwensmall}{Qwen2.5-VL-7B-Instruct}
\newcommand{\corlbl}{Cor+Lbl}
\newcommand{\cor}{Cor}
\newcommand{\iou}{IoU}

\title{\team{} at \shroom{}: Agreement-Guided Ensembling of Zero-Shot and LoRA-Adapted Vision--Language Models}

\author{
  Ali Athar\textsuperscript{1}\quad
  Imran Ahsan\textsuperscript{2} \quad
  Joon-Yong Jung\textsuperscript{1}\footnotemark[1] \\
  \textsuperscript{1}Department of Radiology, Seoul St. Mary's Hospital, The Catholic University of Korea \\
  \textsuperscript{2} Department of Smart City, Chung-Ang University \\
  \texttt{ali.athar1401@gmail.com} \quad
  \texttt{imranahsan23@cau.ac.kr} \quad
  \texttt{messengr@catholic.ac.kr}
}
\begin{document}
\maketitle

\begin{abstract}
This paper describes the SKstars submission to SHROOM-Visions 2026, a shared task on fine-grained hallucination detection in large vision-language model outputs. The task requires systems to identify hallucinated character spans, assign hallucination categories, and provide confidence estimates for their predictions. Our approach combines zero-shot predictions from Qwen2.5-VL-72B-Instruct with those of a LoRA-adapted Qwen2.5-VL-7B-Instruct model. The outputs of the two models are integrated through a lightweight ensemble procedure, followed by span refinement and confidence adjustment. We evaluate the main system components on a small internal development subset and report the performance of the submitted system on the official English test set. SKstars achieved a Cor+Lbl score of 0.2902, ranking 15th among 29 teams, and obtained Cor and IoU scores of 0.3642 and 0.3151, respectively, ranking 18th on both metrics. The results show that combining a large zero-shot model with a smaller adapted model provides a practical framework for multilingual and fine-grained hallucination localization, while also highlighting the difficulty of transferring development-set improvements to hidden test data.
\footnote{Code and predictions: 
\url{https://github.com/aliathar1401/SK-Stars-shroom-visions-2026}}
\end{abstract}

\section{Introduction}

Large vision--language models (LVLMs) have demonstrated 
remarkable capabilities in understanding and describing 
visual content, yet their outputs frequently contain 
claims that are unsupported by, or contradictory to, the 
visual input. Such hallucinations pose significant risks 
in high-stakes applications including medical imaging, 
autonomous systems, and content moderation, where 
factual accuracy is paramount. Existing evaluations often 
reduce this behavior to a response-level or question-level 
decision. For example, POPE focuses on object-existence 
errors \citep{li2023pope}, while HallusionBench evaluates 
image-context reasoning through controlled question pairs 
\citep{guan2024hallusionbench}. These benchmarks are 
useful for comparing models in aggregate, but they do not 
directly identify the exact text that makes an otherwise 
fluent response unreliable, limiting their utility for 
fine-grained error analysis and correction.

The SHROOM series instead treats hallucination detection 
as an annotation problem. The first SHROOM shared task 
studied response-level detection in text generation 
\citep{mickus2024shroom}; Mu-SHROOM later introduced 
multilingual span localization across multiple languages 
\citep{vazquez2025mushroom}. SHROOM-Visions \cite{vazquez2026overview} extends this setting to image-conditioned outputs. A participant 
receives an image, a prompt, and a generated response, 
without knowing which LVLM produced the response. The 
required output marks hallucinated character spans, 
assigns each span one of five categories (invention, 
mischaracterization, OCR problem, miscounting, or other), 
and associates each character with a confidence value 
reflecting the expected level of annotator agreement.

This paper describes our English-track submission under 
the team name \textbf{SKstars}. Our development process 
produced two complementary systems. First, we prompted 
Qwen2.5-VL-72B-Instruct \citep{bai2025qwen25vl} without 
task-specific training, relying on prompt engineering to 
guide hallucination detection. Second, we adapted 
Qwen2.5-VL-7B-Instruct with low-rank updates 
\citep{hu2022lora} on the provided training data. Their 
development IoU scores were nearly identical (0.513 vs. 
0.509 on a 50-example internal diagnostic), but their 
output behavior differed substantially: the larger model 
produced fewer and broader spans and was strongly 
overconfident (98.8\% of spans at confidence 1.0), 
whereas the adapted model generated more spans but 
concentrated most confidence mass at the lowest allowed 
value (93.9\% at 0.33). These complementary failure 
modes motivated our ensemble design: we used the 72B 
spans as anchors and treated cross-model overlap as 
evidence for raising confidence from 0.33 to 0.67.

We report the full pipeline and examine its main failure 
modes. The official ranking is given in the abstract and 
Section~\ref{sec:results}. SKstars achieved a Cor+Lbl 
score of 0.2902 (15th among 29 teams) and Cor and IoU 
scores of 0.3642 and 0.3151, respectively (18th on both 
metrics). The \cor{} score of 0.3642 shows that 
agreement-based confidence was only a partial correction 
to the component models' poorly calibrated outputs, and 
the gap between development IoU (\raise0.4pt\hbox{$\sim$}0.51) 
and official test IoU (0.3151) highlights the difficulty 
of transferring development-set improvements to hidden 
test data containing distribution shifts.
\section{Task and Data}
\label{sec:task}

\paragraph{Task.}
\shroom{} is model-agnostic: systems operate only on the image, prompt, and generated response. They must identify hallucinated character spans and assign one of five labels: \emph{invention} (content absent from the image), \emph{mischaracterization} (visible content described incorrectly), \emph{OCR problem}, \emph{miscounting}, or \emph{other}. The official leaderboard reports character-level intersection-over-union (\iou{}), correlation between predicted and empirical hallucination probabilities (\cor{}), and its label-sensitive counterpart (\corlbl{}). The task specification and submission format are available on the shared-task website.\footnote{\url{https://helsinki-nlp.github.io/shroom/2026}}

\paragraph{Dataset.}
The task data are derived from SHEEP \citep{mickus2026sheep}, a 20,000-item multilingual resource containing English, Chinese, French, and Italian samples. It combines 18,400 outputs from five LVLMs with 1,600 human-written samples and provides fine-grained span annotations from multiple annotators. For English, the task release contains 3,799 labeled training instances and 1,201 test instances. The English test partition includes 400 human-written items in addition to model-generated samples, creating a distribution shift that is not represented in the training partition.

In the English training data, 74.7\% of responses contain at least one aggregated hallucination span, with an average of 4.5 spans among the released annotations. Mischaracterization (44.1\%) and invention (40.4\%) dominate the span labels. Aggregated probabilities take three values reflecting one, two, or three annotators marking a character: 0.33, 0.67, and 1.00. Their training frequencies are 71.2\%, 20.7\%, and 8.1\%, respectively. These statistics motivated our confidence post-processing, although Section~\ref{sec:results} shows why matching a marginal training distribution is not sufficient for calibration.

\section{System Description}
\label{sec:system}

\begin{figure*}[t]
\centering
\resizebox{0.99\textwidth}{!}{%
\begin{tikzpicture}[
  font=\small,
  node distance=7mm and 9mm,
  box/.style={draw, rounded corners=2pt, fill=black!4, align=center,
              minimum height=10mm, text width=31mm, inner sep=4pt},
  smallbox/.style={draw, rounded corners=2pt, fill=black!2, align=center,
              minimum height=9mm, text width=35mm, inner sep=3pt},
  arrow/.style={-{Latex[length=2mm]}, thick}
]
\node[box] (input) {Image, prompt, and\\generated response};
\node[box, above right=5mm and 11mm of input] (large) {Zero-shot\\\qwenlarge{}};
\node[box, below right=5mm and 11mm of input] (small) {LoRA-adapted\\\qwensmall{}};
\node[smallbox, right=15mm of large, yshift=-8mm] (match) {Character-span alignment\\overlap threshold $\geq 0.3$};
\node[smallbox, right=9mm of match] (agree) {72B anchors\\both models: 0.67\\72B only: 0.33\\7B only: discarded};
\node[smallbox, right=9mm of agree] (post) {Remove duplicates\\adjust confidence\\frequencies\\emit task JSON};
\draw[arrow] (input) -- (large);
\draw[arrow] (input) -- (small);
\draw[arrow] (large) -- (match);
\draw[arrow] (small) -- (match);
\draw[arrow] (match) -- (agree);
\draw[arrow] (agree) -- (post);
\end{tikzpicture}%
}
\caption{Overview of the \team{} English submission. The 72B output defines the candidate set; overlap with the adapted 7B model raises confidence from 0.33 to 0.67, but 7B-only spans are discarded.}
\label{fig:pipeline}
\end{figure*}
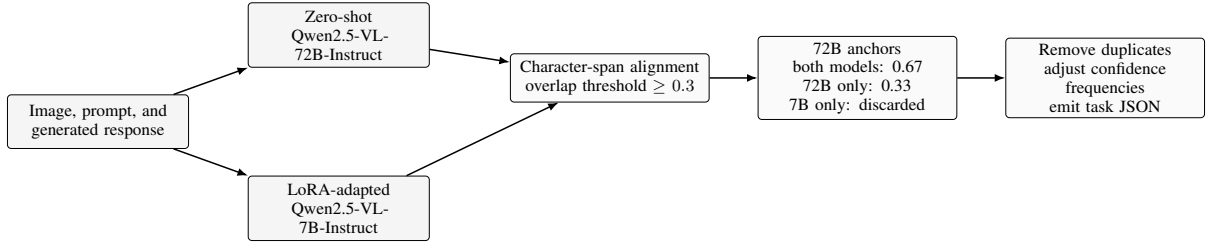

Figure~\ref{fig:pipeline} summarizes the submitted pipeline. Both models receive the same image--prompt--response triplet and are instructed to return literal substrings, labels, and confidence values in JSON. Character offsets are recovered by matching each returned substring to the original response. The final candidate set is seeded from the 72B predictions; the 7B predictions are used to identify cross-model agreement.

\subsection{Zero-Shot 72B Model}

We used \qwenlarge{} \citep{bai2025qwen25vl} with 4-bit NF4 quantization, distributed over three NVIDIA RTX A6000 GPUs with 48~GB memory each. Quantization followed the memory-saving components introduced by QLoRA \citep{dettmers2023qlora} , applied here for inference only.

The prompt served four purposes. It defined the five task labels, required exact copying of hallucinated substrings, requested a strict JSON schema, and reminded the model that hallucinations are common in the task data. The last instruction used the approximate 75\% response-level hallucination rate measured in the training set. This prior increased the model's willingness to flag questionable content, but it also risked pushing predictions toward the training prevalence rather than the evidence in a particular image. We accepted this trade-off given the task's emphasis on recall over precision.

The 72B model reached an \iou{} of 0.513 on our 50-example development diagnostic. Its confidence outputs were unsuitable without post-processing: 98.8\% of predicted spans received 1.0, although only 8.1\% of aggregated training spans had that value. It also returned overlapping or repeated substrings that required deterministic cleanup.

% \subsection{LoRA-Adapted 7B Model}

% We adapted \qwensmall{} with LoRA \citep{hu2022lora}. After reserving data for development and excluding an item whose image was unavailable during training, 3,417 English instances were used for optimization. The adapters had rank $r=16$ and scaling factor $\alpha=32$. They were inserted into the query, key, value, output, gate, up, and down projection matrices, resulting in 47.5 million trainable parameters out of approximately 8.3 billion (0.57\%).

% Images were resized to $448\times448$ pixels to keep multimodal sequence lengths within the available memory. Training used bfloat16 arithmetic, gradient checkpointing, an 8-bit paged AdamW optimizer, a peak learning rate of $2\times10^{-4}$ with cosine decay, and an effective batch size of 8. We retained the one-epoch checkpoint, which required about 80 minutes. Pilot continuation did not yield a consistent improvement in development loss, so later checkpoints were not used in the submitted system.

% The adapted model obtained development \iou{} 0.509, close to the 72B model's 0.513 on the same 50 items. This comparison is encouraging but should be interpreted cautiously: the subset is small, no confidence interval was computed, and the two systems may make different errors despite similar aggregate overlap.

\subsection{LoRA-Adapted 7B Model}
We adapted \qwensmall{} with LoRA \citep{hu2022lora}. 
After reserving data for development and excluding an 
item whose image was unavailable during training, 3,417 
English instances were used for optimization. The 
adapters had rank $r=16$ and scaling factor $\alpha=32$. 
They were inserted into the query, key, value, output, 
gate, up, and down projection matrices, resulting in 
47.5 million trainable parameters out of approximately 
8.3 billion (0.57\%).

Images were resized to $448\times448$ pixels to keep 
multimodal sequence lengths within the available memory. 
Training used bfloat16 arithmetic, gradient 
checkpointing, an 8-bit paged AdamW optimizer, a peak 
learning rate of $2\times10^{-4}$ with cosine decay, 
and an effective batch size of 8. We retained the 
one-epoch checkpoint, which required about 80 minutes. 
Pilot continuation did not yield a consistent 
improvement in development loss, so later checkpoints 
were not used in the submitted system.

The adapted model obtained development \iou{} 0.509, 
close to the 72B model's 0.513 on the same 50 items. 
This comparison is encouraging but should be interpreted 
cautiously: the subset is small, no confidence interval 
was computed, and the two systems may make different 
errors despite similar aggregate overlap. Furthermore, the two models differed substantially in their confidence distributions, motivating the 
ensemble design described next.
% \subsection{Agreement and Post-processing}

% We converted each model's substring predictions to character intervals and matched intervals whose pairwise span \iou{} was at least 0.3. The 72B set formed the backbone of the final output. A 72B span that overlapped a 7B span was assigned an initial confidence of 0.67; an unmatched 72B span received 0.33. Predictions made only by the 7B model were omitted because the 72B model had the slightly higher development \iou{}. This asymmetric rule favored precision and stability over recall.

% We then removed 262 duplicate entries. Before the final redistribution step, the agreement rule yielded 53.6\%, 40.3\%, and 6.1\% of predictions at confidence 0.33, 0.67, and 1.00. For the submitted run, these values were further redistributed toward the training frequencies of 71.2/20.7/8.1. This last operation is a prior-matching heuristic rather than learned calibration: it constrains the global frequency of confidence values but does not guarantee that a particular span receives an accurate probability.

% \subsection{Agreement and Post-processing}
\label{sec:ensemble}
We converted each model's substring predictions 
to character intervals and matched intervals whose 
pairwise span \iou{} was at least 0.3. The 72B set 
formed the backbone of the final output. A 72B span 
that overlapped a 7B span was assigned an initial 
confidence of 0.67; an unmatched 72B span received 
0.33. Predictions made only by the 7B model were 
omitted because the 72B model had the slightly higher 
development \iou{}. This asymmetric rule favored 
precision and stability over recall, though it likely 
discarded genuine hallucination spans that the adapted 
model uniquely identified.

We then removed 262 duplicate entries. Before the 
final redistribution step, the agreement rule yielded 
53.6\%, 40.3\%, and 6.1\% of predictions at confidence 
0.33, 0.67, and 1.00. For the submitted run, these 
values were further redistributed toward the training 
frequencies of 71.2/20.7/8.1. This last operation is 
a prior-matching heuristic rather than learned 
calibration: it constrains the global frequency of 
confidence values but does not guarantee that a 
particular span receives an accurate probability.
% \section{Experimental Setup}
% \label{sec:setup}

% The implementation used PyTorch, Hugging Face Transformers, PEFT, and BitsAndBytes. Inference with the quantized 72B model took approximately five hours for the 1,201 English test instances (about 15 seconds per item). Inference with the adapted 7B model took approximately seven hours under our generation settings. Span alignment, duplicate removal, and file generation required less than one minute on CPU.

% Our internal \iou{} diagnostic used a fixed set of 50 labeled English examples. It was intended for rapid comparison during system construction, not for model selection with statistical confidence. All official scores in Table~\ref{tab:official} are from the final hidden test evaluation and supersede the development numbers.

\section{Experimental Setup}
\label{sec:setup}
The implementation used PyTorch, Hugging Face 
Transformers, PEFT, and BitsAndBytes. Inference 
with the quantized 72B model took approximately 
five hours for the 1,201 English test instances 
(about 15 seconds per item). Inference with the 
adapted 7B model took approximately seven hours 
under our generation settings. Span alignment, 
duplicate removal, and file generation required 
less than one minute on CPU. All experiments 
were run on three NVIDIA RTX A6000 GPUs 
(48\,GB each, 144\,GB total).

Our internal \iou{} diagnostic used a fixed set 
of 50 labeled English examples. It was intended 
for rapid comparison during system construction, 
not for model selection with statistical confidence. 
All official scores in Table~\ref{tab:official} 
are from the final hidden test evaluation and 
supersede the development numbers.

\section{Results and Discussion}
\label{sec:results}
\begin{table}[t]
\centering
\small
\begin{tabular}{lcc}
\toprule
\textbf{Official metric} & \textbf{Score} & \textbf{EN rank} \\
\midrule
\corlbl{} & 0.2902 & 15th \\
\cor{}    & 0.3642 & 18th \\
\iou{}    & 0.3151 & 18th \\
\bottomrule
\end{tabular}
\caption{Official English leaderboard result for \team{}.}
\label{tab:official}
\end{table}
Table~\ref{tab:official} reports the final English 
result. The gap between our development \iou{} 
(roughly 0.51 for each component) and official 
\iou{} (0.3151 for the submitted ensemble) is large. 
Several factors may contribute. The internal 
diagnostic contains only 50 examples and therefore 
has high sampling variance. The submitted system 
also uses asymmetric filtering and confidence-oriented 
post-processing rather than simply taking either 
component model's spans. Finally, the hidden test 
set contains human-written examples that differ 
from the LVLM-generated training data 
\citep{mickus2026sheep}.

\begin{table}[t]
\centering
\small
\setlength{\tabcolsep}{4.5pt}
\begin{tabular}{lccc}
\toprule
\textbf{System} & \textbf{Dev} & \textbf{Test hall.} & 
\textbf{Test spans} \\
 & \textbf{\iou{}} & \textbf{responses} & 
 \textbf{/ response} \\
\midrule
72B zero-shot      & 0.513 & 83.3\% & 1.4 \\
7B + LoRA          & 0.509 & 68.0\% & 2.4 \\
Submitted ensemble & --    & 83.3\% & 1.4 \\
\bottomrule
\end{tabular}
\caption{Development overlap and test-output 
characteristics. Test hallucination rates are system 
predictions, not gold statistics.}
\label{tab:diagnostics}
\end{table}
Table~\ref{tab:diagnostics} shows that similar 
development \iou{} values concealed different 
prediction patterns. The 72B model marked 83.3\% 
of test responses but produced only 1.4 spans per 
response. The adapted model marked fewer responses 
(68.0\%) and produced 2.4 spans per response. 
Relative to the 4.5-span training average, both 
systems tended to merge or omit fine-grained 
annotations, particularly the 72B model. Since the 
ensemble retained only 72B anchors, it inherited the 
larger model's response coverage and span count. 
Discarding all 7B-only spans likely removed useful 
fine-grained candidates together with false positives.

\begin{table}[t]
\centering
\small
\setlength{\tabcolsep}{4.5pt}
\begin{tabular}{lrrr}
\toprule
\textbf{Source} & \textbf{0.33} & \textbf{0.67} & 
\textbf{1.00} \\
\midrule
Training annotations        & 71.2 & 20.7 & 8.1  \\
72B raw                     & 0.0  & 1.2  & 98.8 \\
7B raw                      & 93.9 & 5.3  & 0.8  \\
Agreement output$^{\dagger}$& 53.6 & 40.3 & 6.1  \\
\bottomrule
\end{tabular}
\caption{Percentage of spans at each confidence value. 
$^{\dagger}$Before the final prior-matching 
redistribution.}
\label{tab:confidence}
\end{table}
The confidence distributions in 
Table~\ref{tab:confidence} explain the motivation 
for ensembling. The component models failed in 
opposite directions: the 72B model placed almost 
every prediction at 1.0, whereas the adapted 7B 
model placed almost every prediction at 0.33. 
Agreement produced a less degenerate distribution. 
Nevertheless, the official \cor{} of 0.3642 shows 
that global frequency adjustment did not solve 
instance-level calibration. A better approach would 
learn confidence from held-out examples using 
features such as model agreement, overlap strength, 
label agreement, span length, and image--text 
entailment, while preserving a strictly separated 
calibration set.

Label-aware performance was also weaker than 
unlabeled correlation: \corlbl{} (0.2902) was below 
\cor{} (0.3642). Our ensemble used the 72B prediction 
as the anchor when models overlapped, so agreement 
raised confidence even when the models proposed 
different labels or boundaries. Explicit 
label-consistency checks, category-specific 
thresholds, and retention of high-confidence 7B-only 
spans are direct avenues for improvement.
% \section{Conclusion}

% We presented the \team{} English system for \shroom{} 2026. The submission combined zero-shot 72B predictions with a LoRA-adapted 7B verifier and used character-span overlap to assign confidence. The official run ranked 14th on \corlbl{} and 17th on both \cor{} and \iou{}. Our development results show that a parameter-efficiently adapted 7B model can approach the span overlap of a much larger zero-shot model on a small diagnostic set. The official evaluation, however, makes the main weakness clear: anchoring all output spans to the 72B model limited granularity, and matching marginal confidence frequencies did not provide reliable per-span calibration.

\section{Conclusion }

We presented the SKstars English system for 
SHROOM-Visions 2026. The submission combined 
zero-shot 72B predictions with a LoRA-adapted 
7B verifier and used character-span overlap to 
assign confidence. The official system ranked 
15th on Cor+Lbl and 18th on both Cor and \iou{}. 
Our development results show that a 
parameter-efficiently adapted 7B model can approach 
the span overlap of a much larger zero-shot model 
on a small diagnostic set. The official evaluation, 
however, makes the main weakness clear: anchoring 
all output spans to the 72B model limited 
granularity, and matching marginal confidence 
frequencies did not provide reliable per-span 
calibration.

\section*{Limitations}

The internal comparison is based on only 50 labeled examples and does not include repeated splits, random seeds, uncertainty estimates, or category-level evaluation. It should therefore be read as a development diagnostic rather than evidence that the 7B and 72B models are generally equivalent.

The final ensemble is asymmetric. It discards every 7B-only span and retains the 72B boundary and label for matched predictions. This design may improve precision, but it prevents the adapted model from correcting coarse span boundaries or recovering hallucinations missed by the larger model. The overlap threshold of 0.3 was selected heuristically and was not tuned through a systematic sweep.

Confidence redistribution matches a marginal training distribution rather than estimating conditional correctness. It assumes that the prevalence and ambiguity of hallucinations are stable between training and test data. This assumption is especially questionable because the hidden test split includes human-written samples that are absent from the training split. The method can therefore appear distributionally plausible while assigning poor probabilities to individual characters.

All experiments target English. The approach was not evaluated on Chinese, French, or Italian, so its behavior under different scripts, tokenization patterns, and multilingual OCR conditions is unknown. Resizing every image to $448\times448$ may remove details required for OCR and counting. Finally, the 72B component requires three 48~GB GPUs and several hours of inference, which limits reproducibility for teams with modest hardware.

\section*{Ethical Considerations}

The system is a detector and does not generate new benchmark responses. Its predictions should not be treated as guarantees of factual correctness: false negatives can leave unsupported claims unmarked, while false positives can incorrectly challenge valid content. These risks are most consequential in high-stakes settings, where automatic span detection should support rather than replace human review. The work uses the shared-task data under the terms supplied by the organizers and does not introduce additional personal data.

%\clearpage

\section*{Acknowledgement}
This research was supported by a grant of the Korea Health Technology R\&D Project through the Korea Health Industry Development Institute (KHIDI), funded by the Ministry of Health \& Welfare, Republic of Korea (grant number: RS-2026-25540297), and the Research Fund of Seoul St. Mary’s Hospital, The Catholic University of Korea.

\end{document}